\documentclass[sigconf]{acmart}
\usepackage{adjustbox}
\usepackage{graphicx}
\usepackage{subcaption}
\usepackage{multirow}
\usepackage{tabularx}
\usepackage{threeparttable}
\usepackage{color}
\usepackage[ruled,vlined,linesnumbered]{algorithm2e}
\usepackage{amsmath}
\SetKwComment{Comment}{$\triangleright$\ }{}
\AtBeginDocument{%
  }
\setcopyright{acmlicensed}
\copyrightyear{2025}
\acmYear{2025}
\acmDOI{XXXXXXX.XXXXXXX}
\acmConference[CIKM '25]{Proceedings of the 34th ACM International Conference on Information and Knowledge Management}{November 10--14,
  2025}{Seoul, Republic of Korea}
\acmISBN{978-1-4503-XXXX-X/2018/06}

\begin{document}
\title{From Forecast to Action: Uncertainty-Aware UAV Deployment for Ocean Drifter Recovery}

\author{Jingeun Kim}
\affiliation{%
  \institution{Gachon University}
  \country{Republic of Korea}
}
\email{wlsrms27@gachon.ac.kr}

\author{Yong-Hyuk Kim}
\affiliation{%
  \institution{Kwangwoon University}
  \country{Republic of Korea}}
\email{yhdfly@kw.ac.kr}

\author{Yourim Yoon}
\affiliation{%
  \institution{Gachon University}
  \country{Republic of Korea}}
\email{yryoon@gachon.ac.kr}

\renewcommand{\shortauthors}{J. Kim et al.}

\begin{abstract}

We present a novel \textit{predict-then-optimize} framework for maritime search operations that integrates trajectory forecasting with UAV deployment optimization—an end-to-end approach not addressed in prior work. A large language model predicts the drifter’s trajectory, and spatial uncertainty is modeled using Gaussian-based particle sampling. Unlike traditional static deployment methods, we dynamically adapt UAV detection radii based on distance and optimize their placement using  meta-heuristic algorithms. Experiments on real-world data from the Korean coastline demonstrate that our method, particularly the repair mechanism designed for this problem, significantly outperforms the random search baselines. This work introduces a practical and robust integration of trajectory prediction and spatial optimization for intelligent maritime rescue.
\end{abstract}



\maketitle

\section{INTRODUCTION}

Maritime accidents can cause severe damage to both human life and the environment, making rapid response crucial. According to maritime accident statistics, the incidence of such accidents in South Korea remains high \cite{choe2025development}. In response, some studies have focused on predicting the trajectories of ocean drifters \cite{kim2024evolutionary, jurkus2023application, li2023ship}. Other studies have addressed the deployment optimization of unmanned aerial vehicles (UAVs) to detect drifting particles \cite{lee2024memetic, hong2024memetic}.  Although recent advances in machine learning–based trajectory prediction have achieved practical levels of accuracy, these predictions have rarely been effectively integrated into UAV deployment strategies. To the best of our knowledge, no prior work has proposed an end-to-end framework that simultaneously predicts drifting trajectories and optimizes UAV deployment decisions. Although trajectory prediction has been widely studied, its integration into downstream deployment strategies remains limited. Similarly, UAV deployment research has extensively used meta-heuristic algorithms, yet the specific design of cost functions is often insufficiently considered \cite{elmachtoub2022smart}.

While \textit{predict-then-optimize} frameworks are well established for integrating forecasting with decision-making \cite{kotary2023predict, sadana2025survey, vanderschueren2022predict}, most existing studies have focused on relatively small-scale problems, such as the shortest path or portfolio optimization, where exact optimization methods like linear programming \cite{wilder2019melding}, quadratic programming, or mixed-integer programming (MIP) \cite{elmachtoub2022smart} are applicable. However, applying these frameworks to large-scale, dynamic problems such as UAV deployment for maritime search and rescue poses unique challenges. The vast and continuous search space involved in optimizing the placement of multiple UAVs under uncertainty quickly renders exact methods computationally intractable, making them impractical for time-critical search and rescue (SAR) operations. Consequently, meta-heuristic algorithms are required to efficiently explore the enormous solution space and obtain high-quality solutions.

In this study, we propose a novel \textit{predict-then-optimize} framework that integrates ocean drifter trajectory prediction with UAV deployment optimization, enabling robust and intelligent maritime rescue operations. We employ a language model to forecast drifter trajectories and use simulated annealing (SA), particle swarm optimization (PSO), and genetic algorithm (GA) to optimize UAV deployment around Gaussian-based particle samples near the predicted paths, accounting for potential prediction errors. To evaluate UAV deployment effectiveness, we introduce a custom metric. While prior studies have demonstrated the reliability of language model-based trajectory prediction \cite{kim2024evolutionary}, our work emphasizes the integration of these predictions with an optimized UAV deployment strategy. The framework is validated using real-world oceanographic data.  Our main contributions can be summarized as follows:
 
\begin{itemize}
    \item We present the first end-to-end \textit{predict-then-optimize} framework tailored for a real-world application, a domain previously unexplored by similar frameworks.
    \item We propose a novel repair mechanism designed to enhance the performance of the meta-heuristic algorithms while rigorously enforcing operational constraints.
    \item We introduce an uncertainty-aware problem formulation in which the search area is adaptively scaled based on prediction based on the language model, and individual UAV search capabilities are modeled according to their operational constraints.
\end{itemize}

\section{PROPOSED APPROACH}
\begin{figure}[t!]
    \centering
    \includegraphics[width=0.5\textwidth]{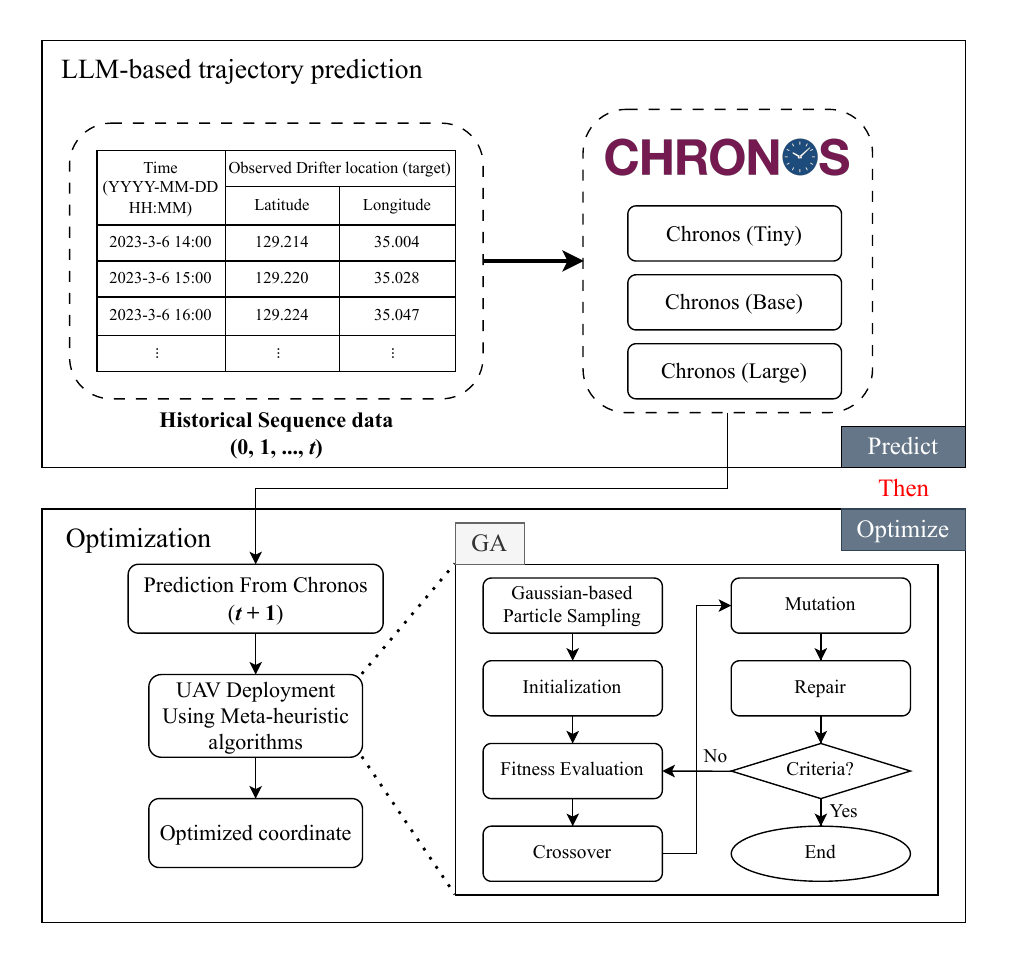}
    \caption{The proposed framework combining LLM-based trajectory prediction with UAV deployment optimization via meta-heuristic algorithms}
    \label{fig:framework}
\end{figure}

 The proposed framework for optimizing UAV deployment is illustrated in Figure \ref{fig:framework}. The process is divided into two main stages: (1) drifter trajectory prediction using a Large Language Model (LLM), and (2) UAV deployment optimization using a SA, PSO, and GA.

 In the first stage, we use Chronos, a pretrained time-series forecasting model, to predict the target's future location at time $(t+1)$ based on its historical sequence data. This predicted coordinate becomes the central point for the subsequent search operation. In the second stage, the optimization phase, SA, PSO, and GA are employed to determine the optimal deployment coordinates for the UAVs. This stage aims to maximize the search coverage for a set of virtual drifter particles generated from a Gaussian distribution centered on the predicted location.

\subsection{Problem Definition}
 The core challenge is to determine the optimal configuration of Unmanned Aerial Vehicles (UAVs) to maximize the likelihood of detecting a drifting object under uncertainty. Formally, the problem can be defined as follows:
 
\begin{itemize}
    \item Given: 
    \begin{itemize}
        \item A set of $n$ available UAVs, $U = \{u_1, u_2, \ldots, u_n\}$.
        \item An operational UAV center point for the search, defined by the language model's predicted drifter location at time $t_p$.
        \item A set of $k$ candidate drifter trajectories, represented as straight lines from the initial accident location to $k$ particle locations sampled from a Gaussian distribution centered at the predicted location.
    \end{itemize}
    \item Decision Variables: The set of 2D deployment coordinates for each UAV, $C = \{(x_1, y_1), (x_2, y_2), \ldots, (x_n, y_n)\}$.
    \item Objective: To find the optimal set of coordinates $C^*$ that maximizes a fitness function that measures the overall coverage of the candidate drifter trajectories, while satisfying constraints on UAV placement and preventing UAV overlap.
\end{itemize}

The following sections detail the models and assumptions used to construct this optimization problem.

\subsection{Problem Modeling}

\subsubsection{Search Area Definition}
The feasible area for UAV deployment, denoted by $A_u$, is defined as a circle centered around the prediction based on the language model at time $t_p$. We refer to this center as the UAV center point, as it serves as a dynamic datum to focus search resources on the most probable location of the drifter. 

The radius $R$ of this area is set to four times the prediction error observed at the previous time step, $t_{p-1}$. This conservative heuristic choice accounts for prediction uncertainty and ensures that the search area is sufficiently large to likely encompass the actual drifter location.

\subsubsection{Dynamic Detection Radius}
\begin{figure}[t]
\centering
\begin{subfigure}[b]{0.35\linewidth}
\centering
\includegraphics[width=\linewidth]{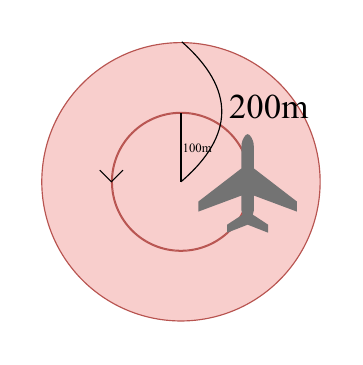}
\caption{}
\label{fig:illuleft}
\end{subfigure}
\hfill
\begin{subfigure}[b]{0.35\linewidth}
\centering
\includegraphics[width=\linewidth]{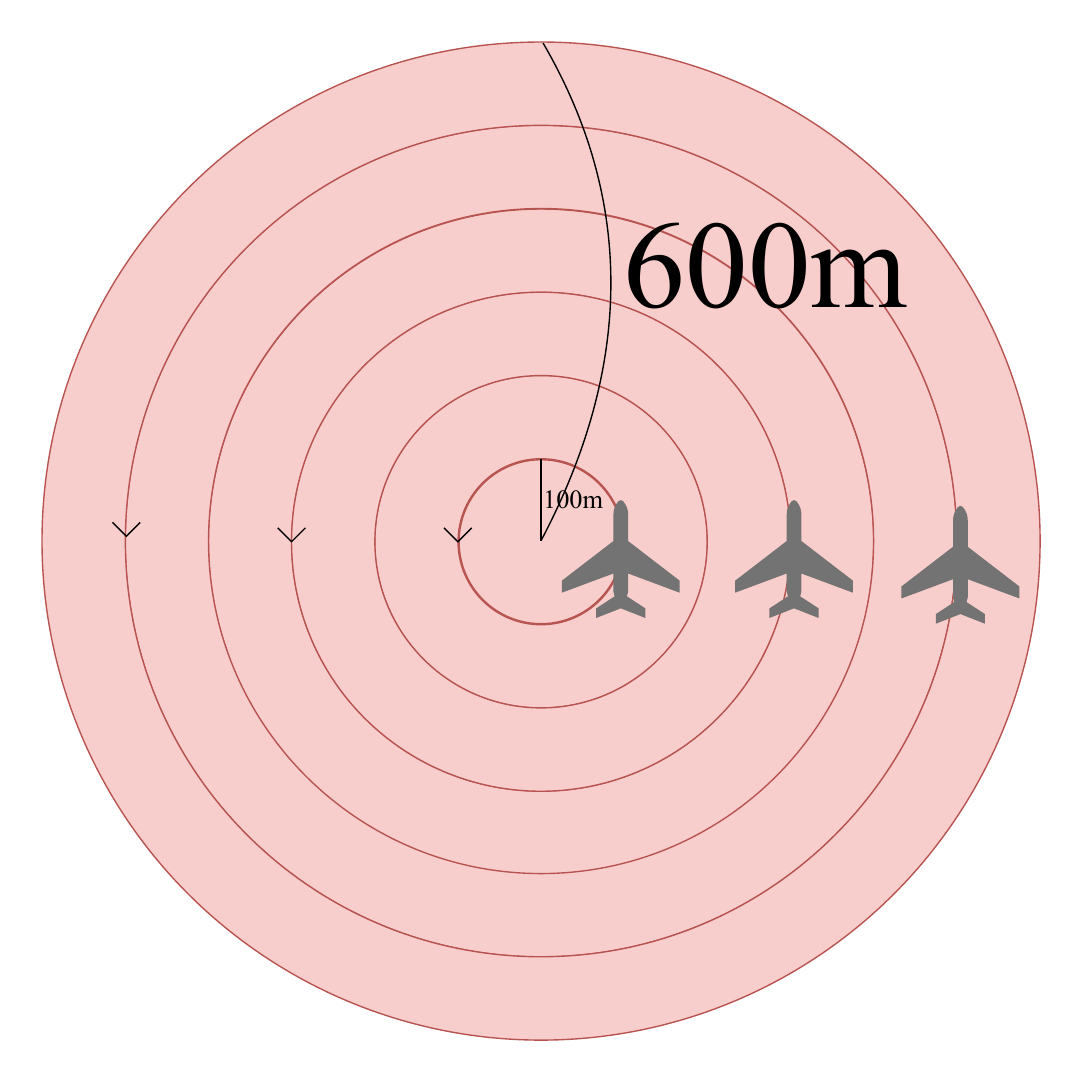}
\caption{}
\label{fig:illuright}
\end{subfigure}
\caption{Illustration of the minimum (200m) and maximum (600m) effective detection radii based on the number of spiral search revolutions.}
\label{fig:illu}
\end{figure}

The UAV model used in this study is the DJI Matrice 300 RTK. While its base detection radius is 100 meters, we model an effective search coverage that extends up to 600 meters, assuming the UAV follows a spiral search pattern with up to three revolutions, as illustrated in Figure \ref{fig:illu}. This operational model is constrained by the UAV's maximum flight time of 55 minutes and travel speed of 1.38 km/min, within which it must reach its assigned coordinate, perform the search, and return \cite{hao2024monitoring}.

A critical assumption in our model is that a UAV's effective detection radius is dynamic and depends on its distance from the UAV center point. A longer travel time to a distant coordinate reduces the time available for the spiral search pattern, thus shrinking the effective search radius. As a simplifying assumption, we model this trade-off as a linear relationship. This results in a dynamic detection radius $D_{r_i}(x_i, y_i)$ ranging from a maximum of 600 meters (close to the center) to a minimum of 200 meters (at the edge of the travel range), as shown in Figure~\ref{fig:illu}(\subref{fig:illuleft}) and \ref{fig:illu}(\subref{fig:illuright}). The radius is defined as: 

\begin{equation}
\begin{aligned}
D_{r_i}(x_i, y_i) = -200 \cdot d + 600
\label{dynamic}
\end{aligned}
\end{equation}
where $d$ is the haversine distance in kilometers between the UAV's coordinate $(x_i, y_i)$ and the UAV center point. The distance is computed using the Haversine formula:

\begin{equation}
\begin{aligned}
d = 2E_r \cdot \arcsin\left( \sqrt{ \sin^2\left( \frac{\Delta \phi}{2} \right) + \cos(\phi_1) \cos(\phi_2) \sin^2\left( \frac{\Delta \lambda}{2} \right) } \right)
\end{aligned}
\end{equation}
where $E_r$ is the Earth's radius, $\phi_1$ and $\phi_2$ denote the latitudes of the UAV and the center point, respectively, and $\lambda_1$ and $\lambda_2$ denote their longitudes. The differences are defined as $\Delta\phi = \phi_2 - \phi_1$ and $\Delta\lambda = \lambda_2 - \lambda_1$.

\subsubsection{Probability of Detection (PoD) Model}
While the geometric coverage is used as the fitness function during optimization for computational efficiency, the final evaluation of a deployment strategy considers the probability of detection. Since the UAV performs a spiral motion in a specific direction, and the ocean drifter may approach from the opposite direction, detection is not guaranteed. Following established SAR literature \cite{guard2002us, agbissoh2019decision}, the probability of detection (PoD) is modeled as the ratio of the dynamic detection radii to the maximum possible detection radii, expressed as:

\begin{align}
PoD &= 1 - e^{-c}
\end{align}
where \( c \) is the coverage factor, defined as the ratio between the $i$-th UAV's effective detection radii $D_{r_i}(x_i, y_i)$ and the maximum detection radii (600m).

\subsection{Prediction}
We used language model to predict the trajectory of an ocean drifter. While previous studies \cite{haldar2025wide, bayindir2024lagrangian, kim2024evolutionary} have demonstrated the effectiveness of various machine learning models, recent advancements in sequence modeling have highlighted the superior performance of transformer-based architectures in capturing long-term dependencies in time series data \cite{su2025systematic, kang2024transformer, xu2024survey}.

 For the prediction part of our \textit{predict-then-optimize} framework, we therefore leverage a large language model (LLM) to estimate the future position of the ocean drifter, $\hat{y}_t$, at time step $t$. Specifically, we employ Chronos, a pretrained foundation model that reframes time series forecasting as a language modeling problem \cite{ansari2024chronos}. This approach allows the model to leverage knowledge learned from a vast and diverse corpus of time series data for robust zero-shot prediction \cite{ansari2024chronos, wang2024deep}.  This zero-shot prediction capability is particularly crucial in real-world rescue scenarios, where the time-critical environment does not allow for online model training.

 The input to our model is a historical sequence of the drifter's coordinates over a context window of length $C$, i.e., $\{y_{t-C}, \dots, y_{t-1}\}$, where each $y_i = (\text{lat}_i, \text{lon}_i)$. Since Chronos is an inherently univariate model, we treat the latitude and longitude as two independent time series, training separate tokenization schemes and making predictions for each. To extend these one-step forecasts into multi-step trajectory prediction, we adopt an iterative forecasting strategy.

 Our prediction strategy follows an iterative forecasting strategy, which is known as a one-step-ahead recursive approach with an expanding window \cite{dong2024temporal}. At each time step, the model predicts only the next value, and the prediction is then fed back as part of the input for subsequent forecasts. Specifically, for the first prediction at time step $t=1$, the model uses the initial coordinate $y_0$ to forecast $\hat{y}_1$. To predict the coordinate at $t=2$, the model is provided with the updated historical sequence $\{y_0, y_1\}$. This process is repeated, so that the prediction at any future time step $t_p$ relies on all available past observations $\{y_0, y_1, \dots, y_{p-1}\}$. 

However, such an iterative forecasting strategy may accumulate errors over time. Even seemingly minor deviations can result in large spatial discrepancies in practice. Notably, in the real world, even a seemingly small difference—such as 0.1 degrees in latitude or longitude, which might appear insignificant in evaluation metrics such as mean squared error (MSE) can translate into several kilometers of physical distance. To address this, we introduced Gaussian-based particle sampling centered around the LLM predictions to better reflect spatial uncertainty and the real-world implications of prediction errors. Spatial uncertainty was modeled by sampling $k$ particles from a bivariate Gaussian distribution centered at the predicted position $\hat{y}_t$. In \cite{brettin2025uncertainty}, the prediction uncertainty in oceanographic applications is often assumed to follow a Gaussian distribution. The covariance matrix was defined as $\Sigma = \sigma^2 I$, where $\sigma$ was set to four times the Haversine distance between the predicted and actual positions at time $t_{p-1}$. Each particle $p_k$ represents a possible location of the ocean drifter, sampled as $p_k \sim \mathcal{N}(\hat{y}_t, \Sigma)$.

\subsection{Optimization}

To facilitate effective UAV deployment along the ocean drifter’s trajectory for detection, Gaussian-based particle samples were generated around the positions predicted by Chronos, a language model-based predictor. We approximated the likely drifter paths by drawing straight lines from the accident location to the Gaussian-sampled positions. The goal of the optimization phase is to determine a UAV deployment configuration that maximizes the number of these lines intersecting the UAV detection radii. Furthermore, the deployment must satisfy constraints that minimize overlap between UAV detection radii and prevent interference among UAVs.In this study, we employed the GA and PSO to optimize the deployment of UAVs.

\subsubsection{Initialization}

As described above, the deployment of UAVs is optimized using a GA. A population of size $N$ is generated, where each individual $u = \{(x_1, y_1), (x_2, y_2), \ldots, (x_u, y_u)\}$ represents the coordinates of UAVs $U$. Each UAV position in an individual $u$ is generated within $A_r$ by randomly selecting an angle $\theta \in [0, 2\pi]$ and a distance $h$ from the UAV control center. The coordinates are then computed as $x_i = x_{\text{center}} + h \cdot \cos(\theta), y_i = y_{\text{center}} + h \cdot \sin(\theta)$, where $(x_{\text{center}}, y_{\text{center}})$ is the coordinate of the UAV control center. Based on Equation~\ref{dynamic}, each UAV in the individual is assigned its own detection radii.

\subsubsection{Genetic Operator} 

From the $N$, $N/2$ pairs are selected randomly, and genetic operator such as crossover and mutation were applied to generate $N$ offspring \cite{aldana2024moody}. We used the blend crossover (BLX-$\alpha$) operator \cite{ranjan2025threshold}, which generates offspring within an extended interval around the parent genes. Offspring $o = \{o_1, o_2, \ldots, o_n\}$ is generated from parents $x=\{x_1, x_2, \ldots, x_n\}$ and $y = \{y_1, y_2, \ldots, y_n\}$, where each $o_i$ is uniformly randomly chosen from the interval $[\min(x_i, y_i) - \alpha D,\ \max(x_i, y_i) + \alpha D]$, with $D = |x_i - y_i|$ and $\alpha$ is constant.

We adopted a mutation operator that replaces each UAV position with new coordinates randomly sampled within a $A_r$ centered at the UAV control center. For each UAV position in the individual, with a mutation probability, a new position \( (x_i, y_i) \) is generated in the same manner as the initialization process.

\subsubsection{Particle Swarm Optimization}
 
We also implemented particle swarm optimization (PSO), another population-based meta-heuristic inspired by the social behavior of bird flocking or fish schooling \cite{bonyadi2017particle}. A swarm of $N$ particles is initialized, where each particle represents a complete UAV deployment solution $u = \{(x_1, y_1), \allowbreak \ldots,\allowbreak (x_u, y_u)\}$. The initial position of each UAV for every particle is generated randomly within the search area $A_r$, identical to the initialization process described for the GA. Furthermore, each particle is assigned an initial velocity vector, typically set to zero or a small random value \cite{guo2025parameters}.
 
In each iteration, the velocity and position of each particle are updated based on its personal best known position ($p_{\text{best}}$) and the global best known position in the swarm ($g_{\text{best}}$) \cite{guo2025parameters}. The velocity $v_i$ and position $x_i$ of the $i$-th particle are updated as follows:

$$v_{i}(t+1) = w \cdot v_{i}(t) + c_1 \cdot r_1 \cdot (p_{\text{best},i} - x_{i}(t)) + c_2 \cdot r_2 \cdot (g_{\text{best}} - x_{i}(t))$$
$$x_{i}(t+1) = x_{i}(t) + v_{i}(t+1)$$
where $w$ is the inertia weight, $c_1$ and $c_2$ are the cognitive and social coefficients, and $r_1, r_2$ are random numbers in $[0, 1]$.  The velocity update is guided by three components: the particle's current momentum (inertia), its tendency to return to its own best-found location (cognitive component), and its tendency to move toward the swarm's best-found location (social component). After each particle's position is updated, the same repair mechanism detailed in Algorithm~\ref{alg:repair} is applied to ensure the feasibility of the solution.

\subsubsection{Simulated Annealing}
 
Simulated Annealing (SA) is a global optimization algorithm inspired by the physical annealing process, a technique that involves heating and controlled cooling of a material to alter its physical properties \cite{kirkpatrick1983optimization}. SA operates by exploring new solutions in the neighborhood of the current solution \cite{khurshid2024hybrid}. When a better solution is found, the algorithm moves toward it to improve performance. This process is repeated over multiple iterations, where at iteration $k$, the cooling parameter controls the acceptance of new solutions. In the early stages, a high cooling parameter allows the algorithm to accept a wider range of solutions and explore the search space broadly. As the iterations progress, the cooling parameter gradually decreases, enabling a more refined search and guiding the algorithm toward the final solution.

\subsubsection{Repair}

\begin{figure}[t!]
    \centering
    \includegraphics[width=0.45\textwidth]{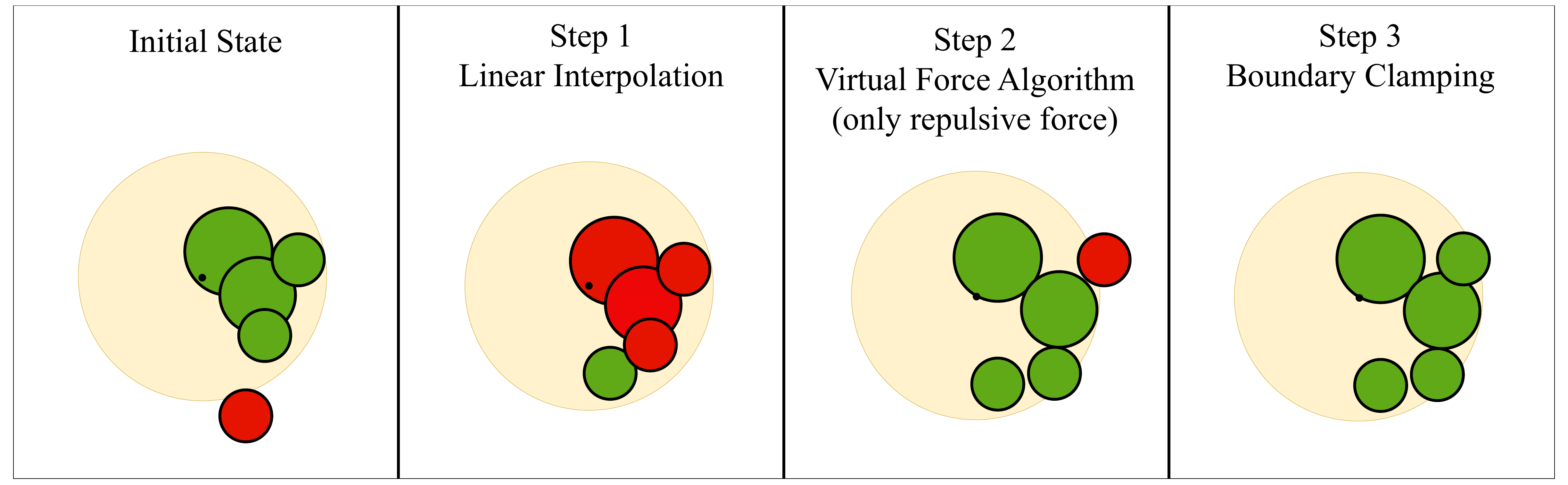}
    \caption{Repair mechanism for UAV deployment}
    \label{fig:repair}
\end{figure}
 
\begin{algorithm}[h!]
\caption{Repair mechanism for UAV Deployment}
\label{alg:repair}
\begin{adjustbox}{max width=\linewidth}
\begin{minipage}{\linewidth}
\SetAlgoLined
\DontPrintSemicolon

\KwIn{
    $P$: Initial UAV positions $\{(lon_1, lat_1), \dots, (lon_N, lat_N)\}$, 
    $R_{uav}$: Detection radii $\{r_1, \dots, r_N\}$, 
    $C_{center}$: Search area center $(lon_c, lat_c)$, 
    $R_{search}$: Search area radius, 
    $max\_iter$: Maximum number of iterations
}

$P_{repaired} \gets P$\;

\textbf{Phase 1: Boundary Correction via Linear Interpolation}\;
\For{$p \gets 1$ \KwTo $N$}{
    $d_c \gets \text{HaversineDistance}(P_{repaired}[p], C_{center})$\;
    \If{$d_c > R_{search}$}{
        $P_{repaired}[p] \gets C_{center} + \frac{P_{repaired}[p] - C_{center}}{||P_{repaired}[p] - C_{center}||} \times R_{search}$\;
    }
}

\textbf{Phase 2-3: Resolve Overlaps using Repulsive Forces and Boundary Clamping}\;
\For{$iter \gets 1$ \KwTo $max\_iter$}{
    $is\_overlapping \gets \text{False}$\;
    $F_{rep} \gets \text{zeros}(N, 2)$ \;
    $n \gets \text{zeros}(N)$

    \For{$i \gets 1$ \KwTo $N$}{
        \For{$j \gets i+1$ \KwTo $N$}{
            $d \gets \text{HaversineDistance}(P_{repaired}[i], P_{repaired}[j])$\;
            $d_{min} \gets R_{uav}[i] + R_{uav}[j]$\;
            \If{$d < d_{min}$}{
                $is\_overlapping \gets \text{True}$\;
                \If{$d > 0$}{
                    $\vec{f} \gets \frac{d_{min}}{d} \cdot (P_{repaired}[j] - P_{repaired}[i])$\;
                    $F_{rep}[i] \gets F_{rep}[i] - \vec{f}$\;
                    $F_{rep}[j] \gets F_{rep}[j] + \vec{f}$\;
                    increment n[i]\;
                    increment n[j]\;
                }
            }
        }
    }

    \If{not $is\_overlapping$}{\textbf{break}}

    \For{$p \gets 1$ \KwTo $N$}{
        $P_{repaired}[p] \gets P_{repaired}[p] + \alpha_r \cdot \frac{F_{rep}[p]}{n[p]}$\;
        $d_c \gets \text{HaversineDistance}(P_{repaired}[p], C_{center})$\;
        \If{$d_c > R_{search}$}{
            $P_{repaired}[p] \gets C_{center} + \frac{P_{repaired}[p] - C_{center}}{||P_{repaired}[p] - C_{center}||} \times R_{search}$\;
        }
    }
}

\Return{$P_{repaired}$}\;
\end{minipage}
\end{adjustbox}
\end{algorithm}

To ensure that UAVs remain within $A_r$ as the generations evolve under the genetic operators, a repair process is required to those that move outside $A_r$. Additionally, since overlapping detection radii between UAVs lead to inefficient search, another repair step is needed to minimize overlaps.  Figure \ref{fig:repair} shows the repair mechanism for UAV deployment. First, it identifies UAVs that do not satisfy the allowed-area constraint (red circle in the Initial State in Figure \ref{fig:repair}) and adjusts their coordinates using linear interpolation to satisfy the boundary constraint (Step 1 in Figure \ref{fig:repair}). Next, the mechanism reduces overlaps between UAVs, as indicated by the red circles in Step 1. After reducing overlaps (green circles in Step 2), it again checks for UAVs violating the constraint (red circle in Step 2) and applies linear interpolation to ensure that all UAVs are positioned within the allowed area (Step 3 in Figure \ref{fig:repair}). The detailed procedures for these repairs are outlined in Algorithm \ref{alg:repair}.

We employed linear interpolation, which estimates intermediate values when the endpoints are known, to relocate UAVs that are outside the $A_r$ area back within the allowed region \cite{kholiavchenko2024kabr, scott2022soccertrack}. Specifically, given a UAV position $p_{repaired}$, the UAV control center $C_{\text{center}}$, and the allowed search area radius $R_{\text{search}}$, the repaired UAV position $p_{\text{repaired}}$ is computed as $p_{\text{repaired}} \leftarrow C_{\text{center}} + \frac{p_{repaired} - C_{\text{center}}}{\|p_{repaired} - C_{\text{center}}\|} \times R_{\text{search}}$. This formula computes the unit vector pointing from the center to the UAV and scales it by the search radius, ensuring that the UAV is placed exactly on the boundary. The same mechanism is used both for the initial correction and as a boundary clamping step after applying repulsive forces.

We then applied the virtual force algorithm (VFA) to minimize overlap between UAVs. In VFA, virtual forces such as attractive and repulsive forces are typically employed to control the movement of agents by drawing them toward targets and pushing them away from obstacles or other agents \cite{yoon2013efficient}. However, in our study, only the repulsive force $\vec{F}_r$ is used to prevent UAVs from overlapping. When the haversine distance between two UAVs, $u_i=(x_i,y_i)$ and $u_j=(x_j,y_j)$, is less than the sum of their detection radii, the positions of $u_i$ and $u_j$ are updated by adding $\alpha_r \cdot \frac{\vec{f}_r}{n_r}$, where $\vec{f}_r \leftarrow \vec{f}_r + \left( \frac{r_{u_i} + r_{u_j}}{d(u_i,u_j)}\right)\cdot (u_j - u_i)$, $d(u_i,u_j)$ is the haversine distance between $u_i$ and $u_j$, $r_{u_i}$ and $r_{u_j}$ are their detection radii, $n_r$ is the number of repulsive interactions, and $\alpha_r$ is a constant weighting parameter.

The application of repulsive forces in VFA can inadvertently push UAVs beyond the boundary of the designated area $A_r$. To rectify such violations, a boundary clamping mechanism is subsequently applied. This final step ensures strict adherence to the spatial constraint, which is prioritized over the minimization of overlaps; any UAV that crosses the boundary is projected back onto the perimeter of $A_r$.

\subsubsection{Evaluation during Meta-heuristic algorithms}
To account for the uncertainty in predictions based on language model, we designated the predicted position as the UAV control center and generated Gaussian-based particle sampling centered around it. These particles were assumed to represent possible coordinates of the ocean drifter at time step $t$. Given the known initial position of the missing ocean drifter, straight lines were drawn from the initial point to each particle. UAV deployment optimization was then performed to maximize coverage of these lines. As the genetic algorithm progressed over generations, each solution was evaluated by discretizing the lines into segments by unit length and calculating the distance between the midpoint of each segment and the center of a UAV. If this distance was smaller than the UAV's detection radii, the segment was considered detected. Based on the number of detected segments, the detection performance was estimated and an approximate fitness score was computed.

\subsubsection{Deployment result evaluation}
We assumed that multiple ocean drifters moving in the same direction should be detected by the UAVs. After the genetic algorithm converges, the following equation is used to evaluate the optimized UAV deployment against the actual trajectory of the ocean drifter:

\begin{equation}
\label{eval}
\begin{aligned}
\text{Coverage} = \lim_{U \to \infty} \left( 
\frac{L(I)}{U} \sum_{i=1}^{U} K_0 \cdot \left(1 - P(D_{i-1})\right)^{i-1} \cdot P(D_i) 
\right)
\end{aligned}
\end{equation}
where $L(I)$ indicates the actual trajectory of missing ocean drifters by UAVs, $U$ is the unit length used to segment $L(\cdot)$, $K_0$ is the number of initial ocean drifter, and $P(D_i)$ and $P(D_{i-1})$ represent the probabilities of detection as defined in Equation \ref{eval}.

\section{EXPERIMENTS}

\subsection{Data sets}

\begin{figure}[t]
    \centering

    \begin{subfigure}[b]{0.45\linewidth}
        \centering
        \includegraphics[width=\linewidth]{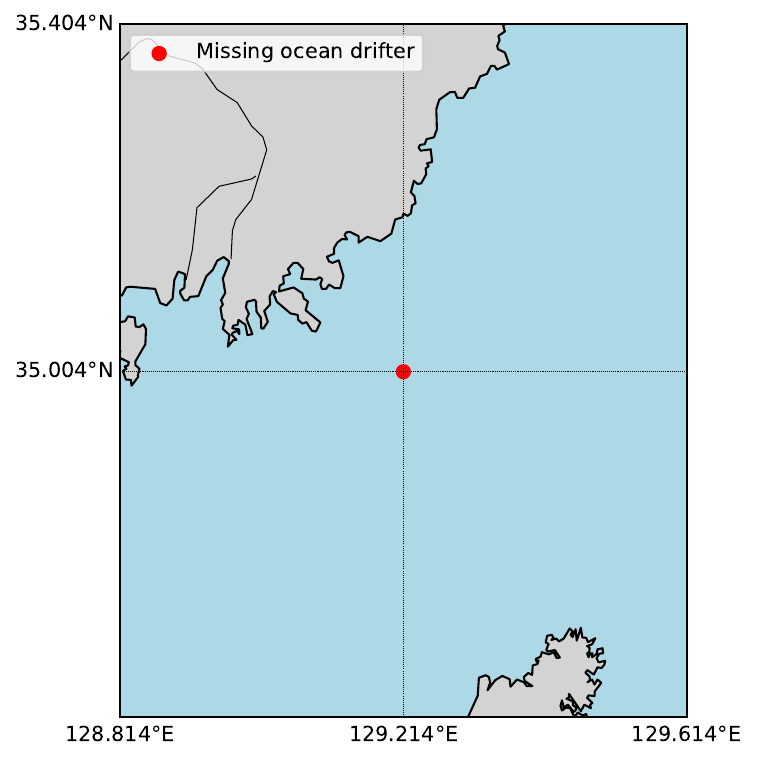}
        \caption{Instance 1}
        \label{fig:img1}
    \end{subfigure}
    \hfill 
    \begin{subfigure}[b]{0.45\linewidth}
        \centering
        \includegraphics[width=\linewidth]{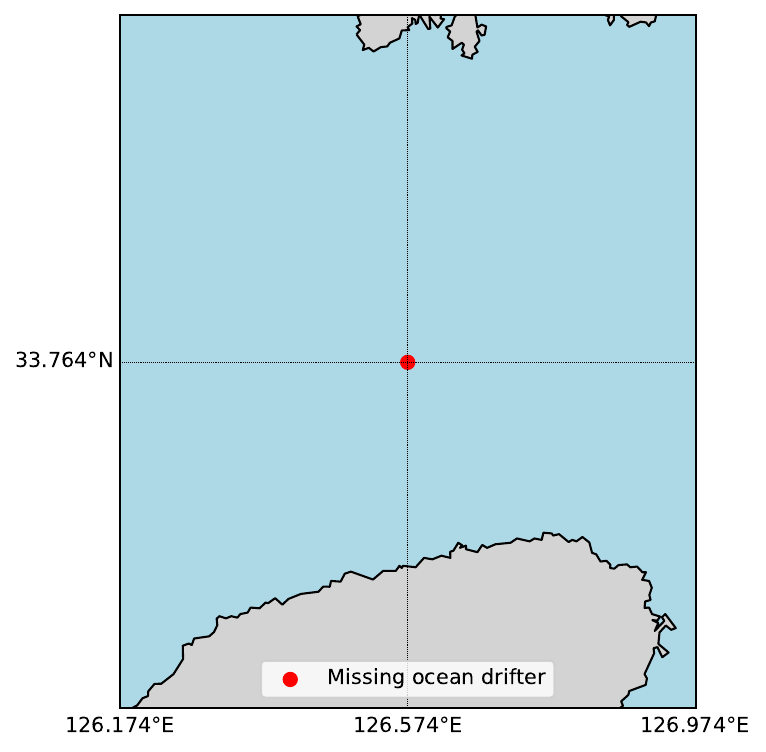}
        \caption{Instance 2}
        \label{fig:img2}
    \end{subfigure}
    
    \vspace{0.5cm} 

    \begin{subfigure}[b]{0.45\linewidth}
        \centering
        \includegraphics[width=\linewidth]{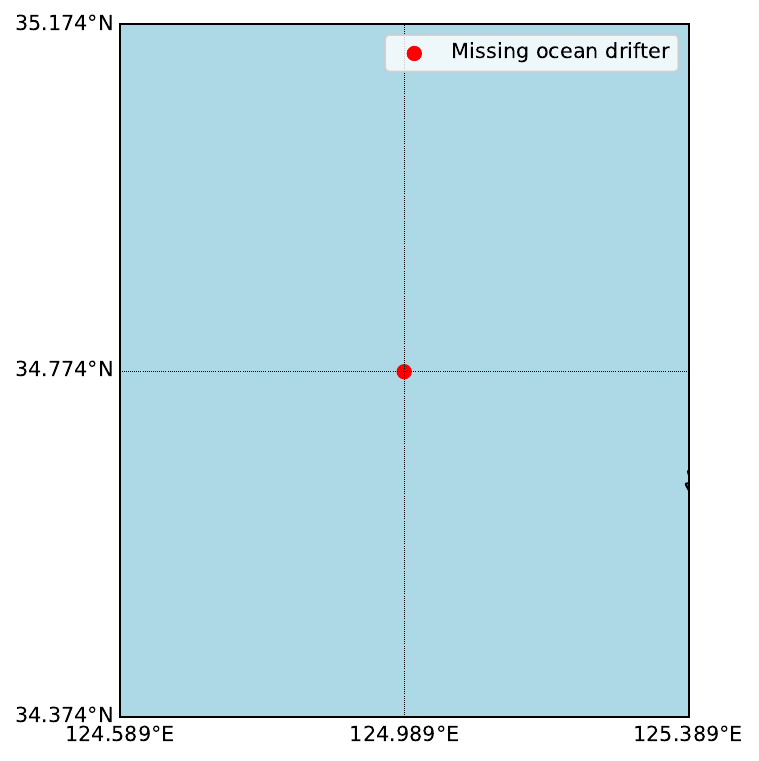} 
        \caption{Instance 3}
        \label{fig:img3}
    \end{subfigure}
    \caption{Visualization of the datasets used in this study. Red dots indicate accident occurrence locations on the map.}
    \label{fig:koreamap}
\end{figure}

In this study, we employed dataset provided by the Korea Hydrographic and Oceanographic Agency. Specifically, we analyzed the trajectories of three ocean drifters deployed in the southern and southwestern coastal region of South Korea. The dataset comprises hourly observations, including the spatial coordinates of wind and ocean current velocities, as well as the latitude and longitude of the drifters. Both wind and ocean current velocities are recorded in meters per second (m/s), and the dataset also provides information regarding the inertial direction of the drifters’ movement. To evaluate the model's performance in predicting the drifter's position at time step \( t_{p+1} \), we employed the subset of data from \( t_2 \) to \( t_p \), within the sequence \( t_1, t_2, \dots, t_p \), as the test set. Five instances were used to evaluate our proposed framework. We assume that the framework is applied six hours after an accident occurs at ocean. Figure \ref{fig:koreamap} presents an overview of the datasets used in this study, where the red point indicates the location of the accident. Figure \ref{fig:koreamap}(\subref{fig:img1}) shows a sample trajectory near the Busan coast, while Figure \ref{fig:koreamap}(\subref{fig:img2}) depicts one near Jeju Island. Figure \ref{fig:koreamap}(\subref{fig:img3}) illustrates a trajectory southwest of Jeju.

\subsection{Experimental setup}

\begin{table}[t]
  \centering
  \caption{Parameter settings for GA, PSO, and SA}
  \label{tab:param_blocks}
  \begin{tabular}{lc}
    \toprule
    \multicolumn{2}{c}{Experimental Settings} \\
    \midrule
    Number of UAVs & 6 and 8 \\
    Number of particles & 10 and 15 \\
    Unit length in optimization & 100 \\
    Unit length at evaluation step & 1 \\
    $K_0$ & 100 \\
    \midrule
    \multicolumn{2}{c}{GA} \\
    \midrule
    Population size & 50 \\
    Number of generations & 50 \\
    Crossover rate & 100\% \\
    Mutation rate & 10\% \\
    $\alpha$ & 0.5 \\
    $\alpha_r$ & 0.9 \\
    \midrule
    \multicolumn{2}{c}{PSO} \\
    \midrule
    Population size & 50 \\
    Number of generations & 50 \\
    Inertia weight $w$ & 0.7 \\
    Cognitive and social coefficients $c_1, c_2$ & 2.0 \\
    \midrule
    \multicolumn{2}{c}{SA} \\
    \midrule
    Initial temperature $T$ & 1.0 \\
    Cooling rate $\alpha$ & 0.95 \\
    Number of iterations & 10 \\
    \bottomrule
  \end{tabular}
\end{table}

Table \ref{tab:param_blocks} lists the settings of parameters used in this study. The population size and the number of generations in the GA were both set to 50. The crossover and mutation rates were set to 100\% and 10\%, respectively. We used three instances to evaluate the proposed framework. For BLX-$\alpha$, the constant parameter $\alpha$ is set to 0.5, which has been widely used \cite{yoon2021maximizing, tsutsui2001search}. In the repair phase of VFA, $\alpha_r$ is set to 0.9. Additionally, different settings—such as the number of particles and UAVs—were applied to assess the generalizability of the framework's performance. During the GA process, a unit length of 100 (equivalent to 100 meters) was used to segment the straight lines between particles and the accident location. In other words, we employed a coarse approximation during the GA to optimize performance efficiently, followed by a detailed evaluation for accuracy After the GA phase, a finer unit length of 1 was used to compare the actual trajectories of the ocean drifters with the UAV deployment results, and the number of initial missing ocean drifters, \( K_0 \), was set to 100.  For PSO, the inertia weight $w$ was set to 0.7, and both cognitive and social coefficients, $c_1$ and $c_2$, were set to 2.0. The SA algorithm was configured with an initial temperature of $T = 1.0$, a cooling rate of $\alpha = 0.95$, and a number of iterations was set to 10.

For trajectory prediction, a language model was built using the Chronos library. We used Python as the implementation language. 

\subsection{Results}

\subsubsection{Prediction}

\begin{table}[t!]
    \centering 
    \caption{Performance comparison of different models using Haversine Distance as the error metric.} 
    \label{tab:model_performance} 
    \begin{tabular}{lr}
        \toprule 
        \textbf{Model} & \textbf{Haversine Distance} \\
        \midrule 
        Random Forest & 1.1263 km \\
        Decision Tree & 1.2339 km \\
        Linear Regression & 1.1427 km \\
        \midrule 
        Chronos-tiny & 0.9793 km \\
        Chronos-base & 1.4780 km \\
        Chronos-large & 1.5112 km \\
        \bottomrule 
    \end{tabular}
\end{table}

To evaluate the predictive performance of the language model, we compared it with models previously employed for ocean drifter forecasting \cite{kim2025evolutionary, kim2024evolutionary}. To mimic real-world scenarios, we calculated the Haversine distance between the predicted and actual values. As shown in Table \ref{tab:model_performance}, while the conventional benchmark models outperformed the larger Chronos-base and Chronos-large variants, which had errors over 1.4 km, the compact Chronos-tiny model achieved superior performance, yielding a lower Haversine error (0.979 km) compared to the best-performing benchmark, Linear Regression (1.142 km).

This superior performance stems from two related factors. The first is the training approach: conventional models are optimized specifically for the ocean dataset, whereas Chronos employs a zero-shot prediction strategy without prior exposure to the data. This provides a crucial advantage in generalizability for new and unseen scenarios. Second, and building on this point, Chronos-tiny's success over its larger siblings is attributed to a better match between model capacity and the domain-specific dataset. Larger pretrained models (base, large) are optimized for broad time-series patterns and may underperform in zero-shot forecasting on small, specialized data where domain characteristics differ substantially from their training corpora. In contrast, the smaller Chronos-tiny model, with lower representational complexity, adapts more robustly to short-term autoregressive patterns without overfitting or amplifying this domain mismatch, making it particularly suitable for this forecasting task.

\subsubsection{Optimization}
\begin{table}[h]
  \centering
  \caption{Comparison of RS, SA, PSO, and GA for different particle and UAV settings. The predictions were produced by the Chronos-tiny model, and Equation \ref{eval} served as the evaluation metric.}
  \label{tab:tiny}
  \resizebox{0.48\textwidth}{!}{%
  \begin{threeparttable}
  \begin{tabular}{@{}cccccccc@{}}
    \toprule
    Instance & \#UAV & \#Particle & Metric & Random & SA & PSO & GA \\
    \midrule
    \multirow{8}{*}{I-1} 
      & \multirow{4}{*}{6} & \multirow{2}{*}{10} & avg  &  30.40 & 54.00 & 46.20 & 41.20 \\
      &   &   & best & 54.00  & 54.00  & 55.00   & 66.00   \\ \cmidrule(lr){3-8}
      &   & \multirow{2}{*}{15} & avg  & 24.40  & 43.80  &  40.20 & 34.00   \\
      &   &   & best & 39.00  & 59.00  & 52.00  & 63.00  \\ \cmidrule(lr){2-8}
      & \multirow{4}{*}{8} & \multirow{2}{*}{10} & avg  & 38.00  & 47.60  &  36.20 & 32.80  \\
      &   &   & best &  39.00 &  65.00 & 61.00  & 54.00  \\ \cmidrule(lr){3-8}
      &   & \multirow{2}{*}{15} & avg  &  34.40 &  49.00 &  49.00 & 45.00  \\
      &   &   & best & 65.00  & 65.00  & 62.00  &  65.00 \\ 
    \midrule
    \multirow{8}{*}{I-2} 
      & \multirow{4}{*}{6} & \multirow{2}{*}{10} & avg  &  43.60 & 40.60  & 43.80 & 48.60 \\
      &   &   & best & 52.00  &  53.00 & 58.00   & 54.00   \\ \cmidrule(lr){3-8}
      &   & \multirow{2}{*}{15} & avg  &  44.80 &  40.40 &  48.80 & 51.20  \\
      &   &   & best & 56.00  & 53.00  & 56.00  & 52.00  \\ \cmidrule(lr){2-8}
      & \multirow{4}{*}{8} & \multirow{2}{*}{10} & avg  & 39.80 & 45.80 & 49.80 & 44.00 \\
      &   &   & best & 52.00 & 51.00 & 52.00 & 53.00 \\ \cmidrule(lr){3-8}
      &   & \multirow{2}{*}{15} & avg  & 49.40 & 48.80 & 47.60 & 52.20 \\
      &   &   & best & 50.00 & 51.00 & 51.00 & 55.00 \\
    \midrule
    \multirow{8}{*}{I-3} 
      & \multirow{4}{*}{6} & \multirow{2}{*}{10} & avg  & 55.60  & 58.80  & 56.00   & 59.80  \\
      &   &   & best & 60.00  & 79.00  & 70.00   & 67.00  \\ \cmidrule(lr){3-8}
      &   & \multirow{2}{*}{15} & avg  & 45.20  & 67.00  & 63.20  & 61.40  \\
      &   &   & best &  62.00 & 78.00  & 74.00  & 68.00  \\ \cmidrule(lr){2-8}
      & \multirow{4}{*}{8} & \multirow{2}{*}{10} & avg  & 67.40  & 67.80  & 66.80  & 61.20  \\
      &   &   & best &  74.00 & 73.00  &  81.00 & 71.00  \\ \cmidrule(lr){3-8}
      &   & \multirow{2}{*}{15} & avg  & 76.60  & 76.60  & 53.00   & 55.80  \\
      &   &   & best & 81.00  & 81.00  & 70.00  &  73.00 \\
    \bottomrule
  \end{tabular}%
  \begin{tablenotes}
    \footnotesize
    \item All methods explored the same number of solutions.
    \item Results are from 5 independent runs with different seeds.
    \item Avg indicates the average performance over the 5 runs.
    \item Best indicates the best performance among the 5 runs.
  \end{tablenotes}
  \end{threeparttable}
  }
\end{table}

\begin{table}[h]
  \centering
  \caption{Comparison of RS, SA, PSO, and GA for different particle and UAV settings. The predictions were produced by the Chronos-base model, and Equation \ref{eval} served as the evaluation metric.}
  \label{tab:base}
  \resizebox{0.48\textwidth}{!}{%
  \begin{threeparttable}
  \begin{tabular}{@{}cccccccc@{}}
    \toprule
    Instance & \#UAV & \#Particle & Metric & Random & SA & PSO & GA \\
    \midrule
    \multirow{8}{*}{I-1} 
      & \multirow{4}{*}{6} & \multirow{2}{*}{10} & avg  & 33.80  &  37.00 & 44.00 & 41.20 \\
      &   &   & best & 44.00  &  49.00 &  60.00  & 51.00   \\ \cmidrule(lr){3-8}
      &   & \multirow{2}{*}{15} & avg  &33.00  & 37.20  &  39.80 &  32.80 \\
      &   &   & best &  41.00 &  59.00 &  55.00 &  61.00 \\ \cmidrule(lr){2-8}
      & \multirow{4}{*}{8} & \multirow{2}{*}{10} & avg  &  37.20 & 44.80  & 43.00  & 33.20  \\
      &   &   & best &  53.00 &  65.00 & 54.00  & 62.00  \\ \cmidrule(lr){3-8}
      &   & \multirow{2}{*}{15} & avg  &  30.00 &  43.20 & 27.20  & 32.40   \\
      &   &   & best &  43.00 &  65.00 & 47.00  & 64.00   \\ 
    \midrule
    \multirow{8}{*}{I-2} 
      & \multirow{4}{*}{6} & \multirow{2}{*}{10} & avg  & 51.40  &  51.80 & 49.20 & 52.20 \\
      &   &   & best &  53.00 &  53.00 & 53.00   & 53.00   \\ \cmidrule(lr){3-8}
      &   & \multirow{2}{*}{15} & avg  &  50.60 &  46.80 & 51.60  & 52.00   \\
      &   &   & best &  53.00 &  53.00 & 53.00  & 53.00   \\ \cmidrule(lr){2-8}
      & \multirow{4}{*}{8} & \multirow{2}{*}{10} & avg  &  51.40 &  51.40 & 50.80  & 52.20   \\
      &   &   & best &  53.00 &  53.00 & 52.00  & 53.00   \\ \cmidrule(lr){3-8}
      &   & \multirow{2}{*}{15} & avg  &  51.00 &  51.00 & 51.00  & 51.00   \\
      &   &   & best &  52.00 &  52.00 & 52.00  & 52.00  \\
    \midrule
    \multirow{8}{*}{I-3} 
      & \multirow{4}{*}{6} & \multirow{2}{*}{10} & avg  &  33.20 &  48.00 & 26.20   & 37.40  \\
      &   &   & best & 42.00  & 69.00  & 56.00   & 49.00 \\ \cmidrule(lr){3-8}
      &   & \multirow{2}{*}{15} & avg &  21.20 &  49.20 & 19.40  & 34.80  \\
      &   &   & best &  33.00 &  69.00 & 34.00  & 38.00   \\ \cmidrule(lr){2-8}
      & \multirow{4}{*}{8} & \multirow{2}{*}{10} & avg  &  38.80 &  44.60 & 44.80  & 21.40   \\
      &   &   & best &  42.00 &  63.00 & 59.00  & 42.00   \\ \cmidrule(lr){3-8}
      &   & \multirow{2}{*}{15} & avg  &  27.40 &  44.80 & 39.20  & 51.00  \\
      &   &   & best &  43.00 &  63.00 & 46.00  & 52.00  \\
    \bottomrule
  \end{tabular}%
  \begin{tablenotes}
    \footnotesize
    \item All methods explored the same number of solutions.
    \item Results are from 5 independent runs with different seeds.
    \item Avg indicates the average performance over the 5 runs.
    \item Best indicates the best performance among the 5 runs.
  \end{tablenotes}
  \end{threeparttable}
  }
\end{table}

\begin{table}[h]
  \centering
  \caption{Comparison of RS, SA, PSO, and GA for different particle and UAV settings. The predictions were produced by the Chronos-large model, and Equation \ref{eval} served as the evaluation metric.}
  \label{tab:large}
  \resizebox{0.48\textwidth}{!}{%
  \begin{threeparttable}
  \begin{tabular}{@{}cccccccc@{}}
    \toprule
    Instance & \#UAV & \#Particle & Metric & Random & SA & PSO & GA \\
    \midrule
    \multirow{8}{*}{I-1} 
      & \multirow{4}{*}{6} & \multirow{2}{*}{10} & avg  & 30.20  & 44.00 & 35.80 & 36.40 \\
      &   &   & best & 48.00  & 47.00  & 54.00   & 52.00   \\ \cmidrule(lr){3-8}
      &   & \multirow{2}{*}{15} & avg  &  35.60 & 38.20  &  46.40 &  32.60 \\
      &   &   & best & 53.00  & 58.00  &  61.00 &  56.00 \\ \cmidrule(lr){2-8}
      & \multirow{4}{*}{8} & \multirow{2}{*}{10} & avg  & 36.20  & 45.60  & 33.00  & 31.60  \\
      &   &   & best &  50.00 & 65.00  & 45.00  & 48.00  \\ \cmidrule(lr){3-8}
      &   & \multirow{2}{*}{15} & avg  &  29.20 & 40.20 & 34.00  & 45.00  \\
      &   &   & best &  42.00 & 65.00  & 52.00  & 65.00  \\ 
    \midrule
    \multirow{8}{*}{I-2} 
      & \multirow{4}{*}{6} & \multirow{2}{*}{10} & avg  &  51.40 & 45.40  & 51.80 & 51.60 \\
      &   &   & best &  54.00 & 52.00  & 55.00   & 54.00   \\ \cmidrule(lr){3-8}
      &   & \multirow{2}{*}{15} & avg  &  51.40 & 48.00  &  46.80 & 51.60  \\
      &   &   & best & 53.00  &  53.00 & 48.00  &  53.00 \\ \cmidrule(lr){2-8}
      & \multirow{4}{*}{8} & \multirow{2}{*}{10} & avg  &  51.80 &  49.00 & 48.20  &  51.60 \\
      &   &   & best &  52.00 & 52.00  & 51.00  & 53.00  \\ \cmidrule(lr){3-8}
      &   & \multirow{2}{*}{15} & avg  &  49.00 & 49.20  &  50.20 & 51.40  \\
      &   &   & best &  51.00 & 51.00  &  53.00 &  53.00 \\
    \midrule
    \multirow{8}{*}{I-3} 
      & \multirow{4}{*}{6} & \multirow{2}{*}{10} & avg  & 35.20  & 46.60  & 29.80   & 52.00  \\
      &   &   & best &  51.00 & 60.00  & 58.00   &  62.00 \\ \cmidrule(lr){3-8}
      &   & \multirow{2}{*}{15} & avg  & 20.60  & 47.20  &  28.80 &  31.20 \\
      &   &   & best & 33.00  &  60.00 &  43.00 &  41.00 \\ \cmidrule(lr){2-8}
      & \multirow{4}{*}{8} & \multirow{2}{*}{10} & avg  &35.40  & 44.20  & 41.20  & 31.20  \\
      &   &   & best & 42.00  & 65.00  & 54.00  & 41.00  \\ \cmidrule(lr){3-8}
      &   & \multirow{2}{*}{15} & avg  & 35.40  & 45.00  & 39.20  & 30.40  \\
      &   &   & best &52.00  & 65.00  & 64.00  & 46.00  \\
    \bottomrule
  \end{tabular}%
  \begin{tablenotes}
    \footnotesize
    \item All methods explored the same number of solutions.
    \item Results are from 5 independent runs with different seeds.
    \item Avg indicates the average performance over the 5 runs.
    \item Best indicates the best performance among the 5 runs.
  \end{tablenotes}
  \end{threeparttable}
  }
\end{table}

 We employed random search (RS), simulated annealing (SA), particle swarm optimization (PSO), and a genetic algorithm (GA) for performance evaluation. The proposed repair mechanism was applied to SA, PSO, and GA, while RS was executed without the repair mechanism. This setup allowed us to comprehensively assess the effectiveness of the repair mechanism across different optimization strategies. The experiment was conducted over 5 runs with different random seeds. The metric indicates how well the deployment of UAVs covers the trajectory of the ocean drifter. 

Tables \ref{tab:tiny}, \ref{tab:base}, and \ref{tab:large} present the results under different particle settings and numbers of UAVs for three instances and different language models. All methods explored the same number of solutions, and the proposed method demonstrated robustness under this fair setting, consistently outperforming the other methods.

\begin{figure}[t]
\subfloat[Average performances]{{\includegraphics[width=0.23\textwidth ]{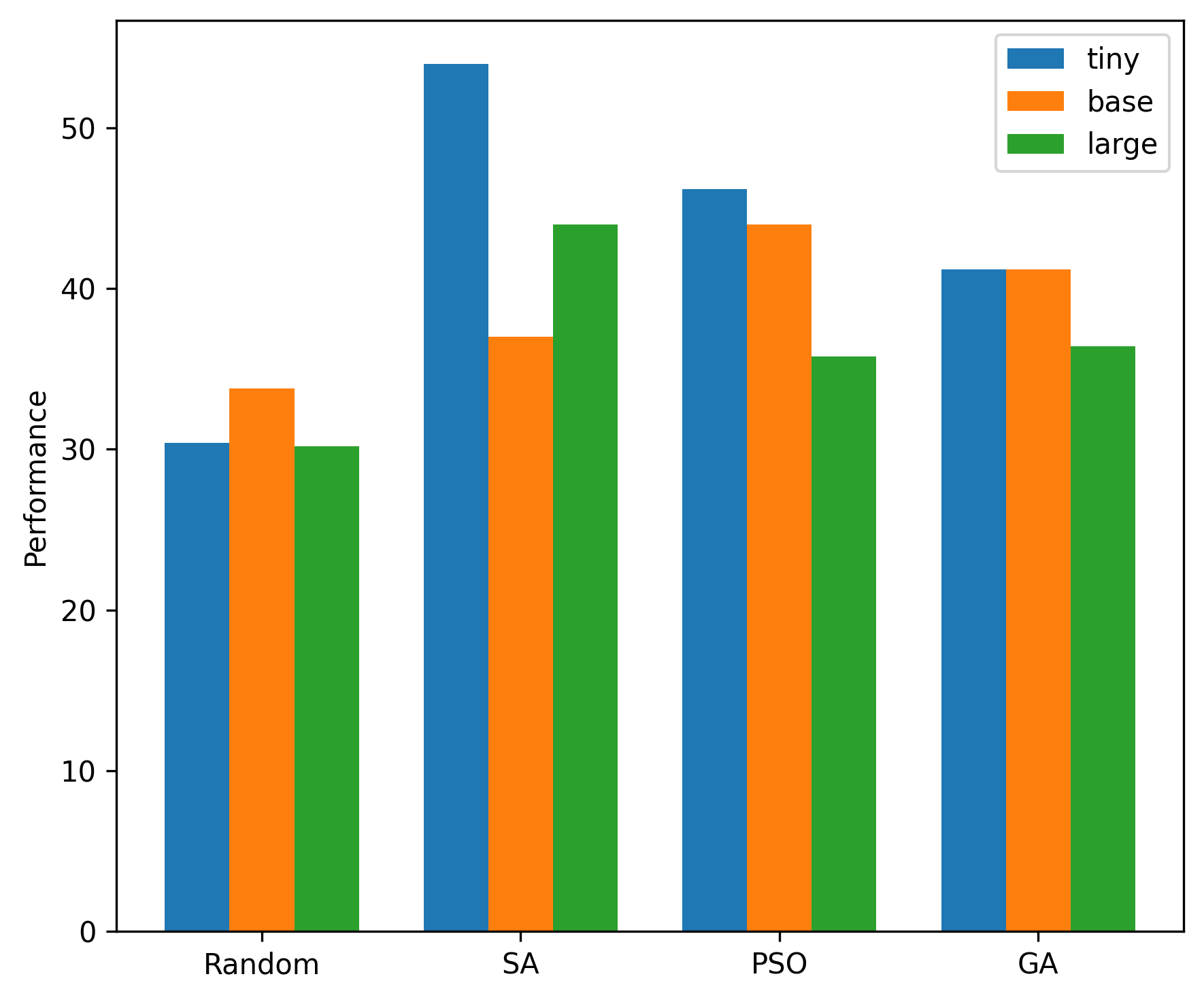} }}%
\subfloat[Best performances]{{\includegraphics[width=0.23\textwidth ]{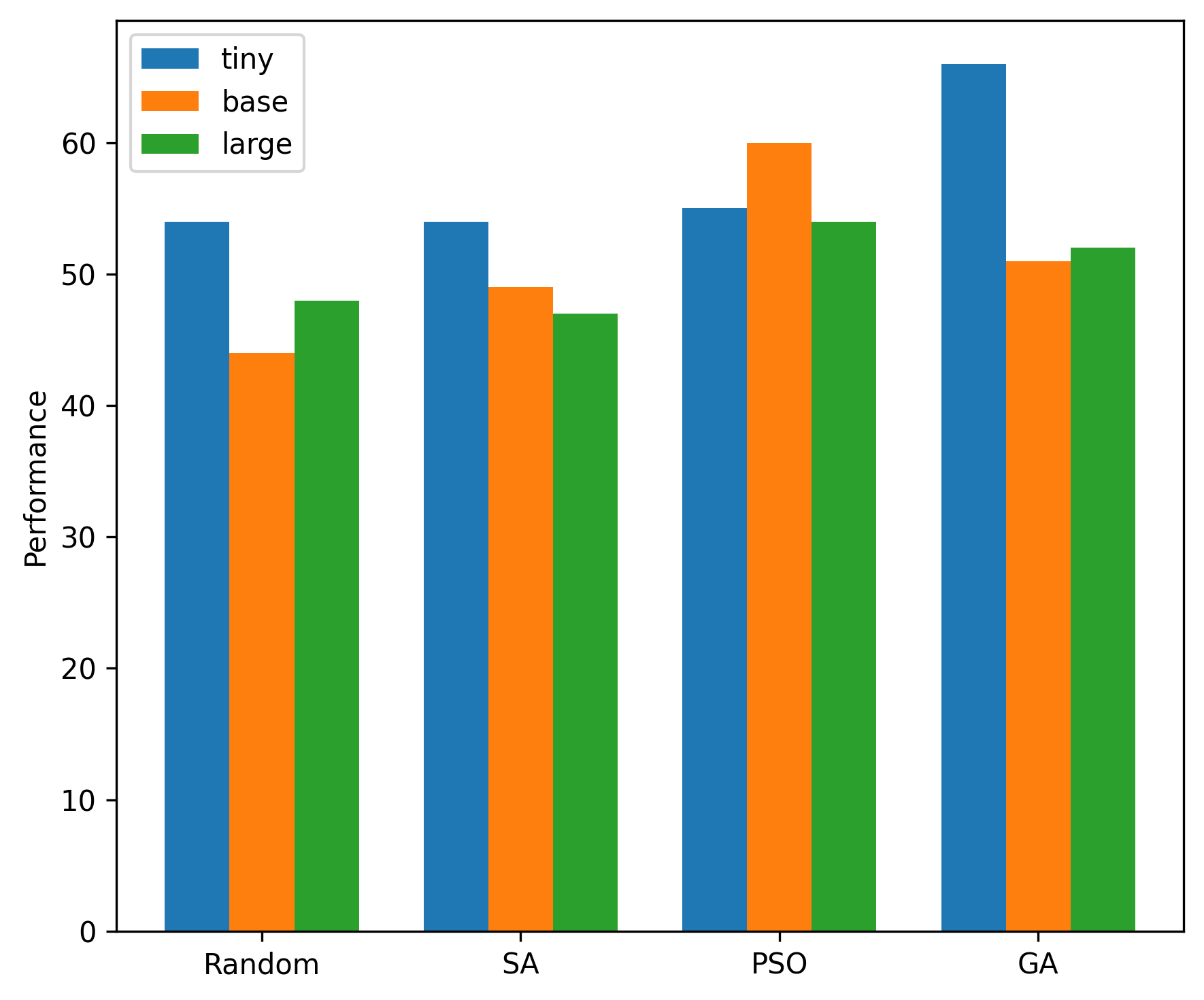} }}%
\caption{Performance comparison of GA, PSO, SA, and random search on Instance 1 with 6 UAVs and 10 particles across Chronos variations}%
\label{fig:result}%
\end{figure}
The results indicate that the meta-heuristic algorithms, SA, PSO, and GA, demonstrated superior and more robust performance compared to RS. The few exceptions to this trend occurred in cases where the resulting solutions featured significant overlap among UAVs, rendering them less cost-effective. Figure \ref{fig:result} visualizes the deployment solutions produced by each method for Instance 1, configured with 6 UAVs and 10 particles using Chronos variations. As shown, RS yielded an average performance of only 30.40, whereas SA, PSO, and GA achieved significantly higher averages of 54.00, 46.20, and 41.20, respectively. Furthermore, in terms of the best solutions, RS reached 54.00, while SA, PSO, and GA attained 54.00, 55.00, and 66.00, respectively, clearly demonstrating the robustness of the meta-heuristic approaches over random search.

\section{Discussion}
The experimental results indicate that the integration of meta-heuristic algorithms into the optimization stage of the predict-then-optimize framework. As demonstrated across Tables \ref{tab:tiny}, \ref{tab:base}, and \ref{tab:large}, the guided search strategies of SA, PSO, and GA consistently achieve better performance than random search, confirming the benefit of guided exploration in this problem setting. However, the relative performance among SA, PSO, and GA varies across instances and parameter configurations, making it difficult to identify a single dominant method. Each algorithm relies on different search dynamics, with SA favoring local refinement, PSO emphasizing swarm-based exploration, and GA maintaining population diversity. As a result, their overall effectiveness is comparable, and the findings suggest that meta-heuristic approaches as a whole provide a more robust solution strategy than unguided random search.

Nevertheless, in a few rare cases, such as Instance 3 with 8 UAVs in Table \ref{tab:tiny}, random search yielded comparable or even better results than GA or PSO. This typically occurred when the solutions generated by the meta-heuristic algorithms resulted in significant overlap among UAV detection ranges, leading to redundant and inefficient coverage. This highlights the importance of overlap handling and the potential benefit of further improving the repair mechanism. Future work could explore adaptive constraint-handling techniques or explicit overlap penalization within the fitness function to strengthen the robustness of GA and PSO across diverse scenarios.

We analyzed the computational complexity of the evaluation metric (Equation \ref{eval}) to assess its feasibility and practicality for real-world applications. The summation requires $U$ iterations, where the most computationally expensive operation in each step is the exponentiation $(1 - P(D_{i-1}))^{i-1}$ with a time complexity of $\mathcal{O}(\log U)$. Here, $U$ denotes the total number of segments dividing the drifter’s trajectory. Consequently, the overall complexity is $\mathcal{O}(U \log U)$. Given this computational cost, we reserve this more precise metric for the final evaluation stage after the optimization process.

\section{CONCLUSION}
We devised a novel \textit{predict-then-optimize} framework that integrates trajectory prediction of a missing ocean drifter with UAV deployment optimization. To address the uncertainty in predictions based on language model, we designed a Gaussian-based particle sampling method. The problem was formulated using real-world data, along with a custom evaluation metric to quantify how well the actual trajectory of the missing drifter was covered. The proposed framework exhibited both strong performance and robustness across multiple case studies.

 Although the proposed framework is demonstrated in a maritime context, it can be readily applied to other spatio-temporal scenarios such as wildfire monitoring, urban traffic control, animal migration tracking, or disaster response, where predictive uncertainty must be accounted for in real-time resource deployment. In these domains, forecasting models can provide short-term trajectory or demand predictions, while optimization components dynamically allocate limited resources such as drones, sensors, or response units. By explicitly incorporating prediction uncertainty into the optimization stage, the framework enables more resilient and adaptive decision-making, ensuring that critical resources are deployed where they are most likely to be effective. This in turn improves both the efficiency and reliability of real-time decision-making. Nonetheless, this study has several limitations that warrant further investigation. Future work should expand the number of test cases and include comparisons with other optimization algorithms to more rigorously evaluate robustness. Additionally, future research should explore various state-of-the-art transformer-based models for time-series forecasting, including models that incorporate exogenous variables as features, to enable a more comprehensive comparison of model performance.

\section{Acknowledgments}
This research was supported by Korea Institute of Marine Science \& Technology Promotion (KIMST) funded by the Ministry of Oceans and Fisheries, Korea (RS-2022-KS221629). This research was also supported by Basic Science Research Program through the
National Research Foundation of Korea(NRF) funded by the Ministry of Education(No. RS-2025-25419851)

\section*{GenAI Usage Disclosure}
This paper was proofread using ChatGPT to improve grammar and language clarity. No content was generated by GenAI.


\bibliographystyle{ACM-Reference-Format}
\bibliography{sample-base}

@article{choe2025development,
  title={Development of Spatial Clustering Method and Probabilistic Prediction Model for Maritime Accidents},
  author={Choe, Cheol-Won and Lim, Suhwan and Kim, Dong Jun and Park, Ho-Chul},
  journal={Applied Ocean Research},
  volume={154},
  pages={104317},
  year={2025},
  publisher={Elsevier}
}

@inproceedings{kim2024evolutionary,
  title={Evolutionary Ensemble for Predicting Drifter Trajectories Based on Genetic Feature Selection},
  author={Kim, Tae-Hoon and Moon, Seung-Hyun and Kim, Yong-Hyuk},
  booktitle={Proceedings of the Genetic and Evolutionary Computation Conference Companion},
  pages={675--678},
  year={2024}
}

@inproceedings{hong2024memetic,
  title={A Memetic Algorithm for Deployment of Search and Rescue Units},
  author={Hong, Seung-Yeol and Kim, Yong-Hyuk},
  booktitle={Proceedings of the Genetic and Evolutionary Computation Conference Companion},
  pages={95--96},
  year={2024}
}

@inproceedings{lee2024memetic,
  title={A Memetic Algorithm to Identify Search Patterns for Maximal Coverage of Drifting Oceanic Particles},
  author={Lee, So-Jung and Kim, Yong-Hyuk},
  booktitle={Proceedings of the Genetic and Evolutionary Computation Conference Companion},
  pages={679--682},
  year={2024}
}

@article{elmachtoub2022smart,
  title={Smart “predict, then optimize”},
  author={Elmachtoub, Adam N and Grigas, Paul},
  journal={Management Science},
  volume={68},
  number={1},
  pages={9--26},
  year={2022},
  publisher={INFORMS}
}

@inproceedings{wilder2019melding,
  title={Melding the data-decisions pipeline: Decision-focused learning for combinatorial optimization},
  author={Wilder, Bryan and Dilkina, Bistra and Tambe, Milind},
  booktitle={Proceedings of the AAAI Conference on Artificial Intelligence},
  volume={33},
  number={01},
  pages={1658--1665},
  year={2019}
}

@article{bayindir2024lagrangian,
  title={Lagrangian Drifter Path Identification and Prediction: SINDy vs Neural ODE},
  author={Bayindir, Cihan and Ozaydin, Fatih and Altintas, Azmi Ali and Eristi, Tayyibe and Alan, Ali Riza},
  journal={arXiv preprint arXiv:2411.04350},
  year={2024}
}

@article{haldar2025wide,
  title={Wide-Range Ocean Current Speed Estimation From Buoy Measurement Data Using Machine Learning},
  author={Haldar, Biswajit and George, Boby and Manickavasagam, Arul Muthiah and Aravindakshan, Atmanand Malayath},
  journal={IEEE Transactions on Instrumentation and Measurement},
  year={2025},
  publisher={IEEE}
}

@article{agbissoh2019decision,
  title={A decision-making algorithm for maritime search and rescue plan},
  author={Agbissoh Otote, Donatien and Li, Benshuai and Ai, Bo and Gao, Song and Xu, Jing and Chen, Xiaoying and Lv, Guannan},
  journal={Sustainability},
  volume={11},
  number={7},
  pages={2084},
  year={2019},
  publisher={MDPI}
}

@article{guard2002us,
  title={US COAST GUARD ADDENDUM},
  author={Guard, Coast},
  year={2002}
}

@article{tsutsui2001search,
  title={Search space boundary extension method in real-coded genetic algorithms},
  author={Tsutsui, Shigeyoshi and Goldberg, Devid E},
  journal={Information Sciences},
  volume={133},
  number={3-4},
  pages={229--247},
  year={2001},
  publisher={Elsevier}
}

@article{yoon2021maximizing,
  title={Maximizing the coverage of sensor deployments using a memetic algorithm and fast coverage estimation},
  author={Yoon, Yourim and Kim, Yong-Hyuk},
  journal={IEEE Transactions on Cybernetics},
  volume={52},
  number={7},
  pages={6531--6542},
  year={2021},
  publisher={IEEE}
}

@article{yoon2013efficient,
  title={An efficient genetic algorithm for maximum coverage deployment in wireless sensor networks},
  author={Yoon, Yourim and Kim, Yong-Hyuk},
  journal={ieee transactions on cybernetics},
  volume={43},
  number={5},
  pages={1473--1483},
  year={2013},
  publisher={IEEE}
}

@article{aldana2024moody,
  title={MOODY: An ontology-driven framework for standardizing multi-objective evolutionary algorithms},
  author={Aldana-Mart{\'\i}n, Jos{\'e} F and del Mar Rold{\'a}n-Garc{\'\i}a, Mar{\'\i}a and Nebro, Antonio J and Aldana-Montes, Jos{\'e} F},
  journal={Information Sciences},
  volume={661},
  pages={120184},
  year={2024},
  publisher={Elsevier}
}

@inproceedings{kholiavchenko2024kabr,
  title={KABR: In-situ dataset for kenyan animal behavior recognition from drone videos},
  author={Kholiavchenko, Maksim and Kline, Jenna and Ramirez, Michelle and Stevens, Sam and Sheets, Alec and Babu, Reshma and Banerji, Namrata and Campolongo, Elizabeth and Thompson, Matthew and Van Tiel, Nina and others},
  booktitle={Proceedings of the IEEE/CVF Winter Conference on Applications of Computer Vision},
  pages={31--40},
  year={2024}
}

@inproceedings{scott2022soccertrack,
  title={SoccerTrack: A dataset and tracking algorithm for soccer with fish-eye and drone videos},
  author={Scott, Atom and Uchida, Ikuma and Onishi, Masaki and Kameda, Yoshinari and Fukui, Kazuhiro and Fujii, Keisuke},
  booktitle={Proceedings of the IEEE/CVF Conference on Computer Vision and Pattern Recognition},
  pages={3569--3579},
  year={2022}
}

@article{li2023ship,
  title={Ship trajectory prediction based on machine learning and deep learning: A systematic review and methods analysis},
  author={Li, Huanhuan and Jiao, Hang and Yang, Zaili},
  journal={Engineering Applications of Artificial Intelligence},
  volume={126},
  pages={107062},
  year={2023},
  publisher={Elsevier}
}

@article{jurkus2023application,
  title={Application of coordinate systems for vessel trajectory prediction improvement using a recurrent neural networks},
  author={Jurkus, Robertas and Venskus, Julius and Treigys, Povilas},
  journal={Engineering Applications of Artificial Intelligence},
  volume={123},
  pages={106448},
  year={2023},
  publisher={Elsevier}
}

@article{vanderschueren2022predict,
  title={Predict-then-optimize or predict-and-optimize? An empirical evaluation of cost-sensitive learning strategies},
  author={Vanderschueren, Toon and Verdonck, Tim and Baesens, Bart and Verbeke, Wouter},
  journal={Information Sciences},
  volume={594},
  pages={400--415},
  year={2022},
  publisher={Elsevier}
}

@article{sadana2025survey,
  title={A survey of contextual optimization methods for decision-making under uncertainty},
  author={Sadana, Utsav and Chenreddy, Abhilash and Delage, Erick and Forel, Alexandre and Frejinger, Emma and Vidal, Thibaut},
  journal={European Journal of Operational Research},
  volume={320},
  number={2},
  pages={271--289},
  year={2025},
  publisher={Elsevier}
}

@article{kotary2023predict,
  title={Predict-then-optimize by proxy: Learning joint models of prediction and optimization},
  author={Kotary, James and Di Vito, Vincenzo and Christopher, Jacob and Van Hentenryck, Pascal and Fioretto, Ferdinando},
  journal={arXiv preprint arXiv:2311.13087},
  year={2023}
}

@article{ranjan2025threshold,
  title={Threshold based constrained $\theta$-NSGA-III algorithm to solve many-objective optimization problems},
  author={Ranjan, Shalu and Gupta, Rachana and Nanda, Satyasai Jagannath},
  journal={Information Sciences},
  volume={697},
  pages={121751},
  year={2025},
  publisher={Elsevier}
}

@article{hao2024monitoring,
  title={Monitoring the spatial--temporal distribution of invasive plant in urban water using deep learning and remote sensing technology},
  author={Hao, Zhenbang and Lin, Lili and Post, Christopher J and Mikhailova, Elena A},
  journal={Ecological Indicators},
  volume={162},
  pages={112061},
  year={2024},
  publisher={Elsevier}
}

@article{ansari2024chronos,
  title={Chronos: Learning the language of time series},
  author={Ansari, Abdul Fatir and Stella, Lorenzo and Turkmen, Caner and Zhang, Xiyuan and Mercado, Pedro and Shen, Huibin and Shchur, Oleksandr and Rangapuram, Syama Sundar and Arango, Sebastian Pineda and Kapoor, Shubham and others},
  journal={arXiv preprint arXiv:2403.07815},
  year={2024}
}

@article{wang2024deep,
  title={Deep time series models: A comprehensive survey and benchmark},
  author={Wang, Yuxuan and Wu, Haixu and Dong, Jiaxiang and Liu, Yong and Long, Mingsheng and Wang, Jianmin},
  journal={arXiv preprint arXiv:2407.13278},
  year={2024}
}

@article{dong2024temporal,
  title={A temporal prediction model for ship maneuvering motion based on multi-head attention mechanism},
  author={Dong, Lei and Wang, Hongdong and Lou, Jiankun},
  journal={Ocean Engineering},
  volume={309},
  pages={118464},
  year={2024},
  publisher={Elsevier}
}

@article{su2025systematic,
  title={A systematic review for transformer-based long-term series forecasting},
  author={Su, Liyilei and Zuo, Xumin and Li, Rui and Wang, Xin and Zhao, Heng and Huang, Bingding},
  journal={Artificial Intelligence Review},
  volume={58},
  number={3},
  pages={80},
  year={2025},
  publisher={Springer}
}

@article{xu2024survey,
  title={Survey and Taxonomy: The Role of Data-Centric AI in Transformer-Based Time Series Forecasting},
  author={Xu, Jingjing and Wu, Caesar and Li, Yuan-Fang and Danoy, Gregoire and Bouvry, Pascal},
  journal={arXiv preprint arXiv:2407.19784},
  year={2024}
}

@article{kang2024transformer,
  title={Transformer-based multivariate time series anomaly detection using inter-variable attention mechanism},
  author={Kang, Hyeongwon and Kang, Pilsung},
  journal={Knowledge-Based Systems},
  volume={290},
  pages={111507},
  year={2024},
  publisher={Elsevier}
}

@article{bonyadi2017particle,
  title={Particle swarm optimization for single objective continuous space problems: a review},
  author={Bonyadi, Mohammad Reza and Michalewicz, Zbigniew},
  journal={Evolutionary computation},
  volume={25},
  number={1},
  pages={1--54},
  year={2017},
  publisher={MIT Press}
}

@article{guo2025parameters,
  title={Parameters identification of magnetorheological damper based on particle swarm optimization algorithm},
  author={Guo, Qianqian and Yang, Xiaolong and Li, Kangjun and Li, Decai},
  journal={Engineering Applications of Artificial Intelligence},
  volume={143},
  pages={110016},
  year={2025},
  publisher={Elsevier}
}

@inproceedings{kim2025evolutionary,
  title={Evolutionary Ensemble for Prediction of Drifter Trajectories Using Weighted Majority},
  author={Kim, Tae-Hoon and Moon, Seung-Hyun and Kim, Yong-Hyuk},
  booktitle={Proceedings of the Genetic and Evolutionary Computation Conference Companion},
  pages={843--846},
  year={2025}
}

@article{khurshid2024hybrid,
  title={A hybrid evolution strategies-simulated annealing algorithm for job shop scheduling problems},
  author={Khurshid, Bilal and Maqsood, Shahid},
  journal={Engineering Applications of Artificial Intelligence},
  volume={133},
  pages={108016},
  year={2024},
  publisher={Elsevier}
}

@article{kirkpatrick1983optimization,
  title={Optimization by simulated annealing},
  author={Kirkpatrick, Scott and Gelatt Jr, C Daniel and Vecchi, Mario P},
  journal={Science},
  volume={220},
  number={4598},
  pages={671--680},
  year={1983},
  publisher={American association for the advancement of science}
}

@article{brettin2025uncertainty,
  title={Uncertainty-permitting machine learning reveals sources of dynamic sea level predictability across daily-to-seasonal timescales},
  author={Brettin, Andrew and Zanna, Laure and Barnes, Elizabeth A},
  journal={arXiv preprint arXiv:2502.11293},
  year={2025}
}

\end{document}